\documentclass{article}

\usepackage{microtype}
\usepackage{graphicx}
\usepackage{subfigure}
\usepackage{booktabs} 
\usepackage{float}
\usepackage{hyperref}

\usepackage[accepted]{icml2025}

\usepackage{amsmath}
\usepackage{amssymb}
\usepackage{mathtools}
\usepackage{amsthm}

\usepackage{algorithm}
\usepackage{algorithmic}
\usepackage{amsmath}
\usepackage{array}
\usepackage{afterpage}
\usepackage{multirow}

\usepackage[capitalize,noabbrev]{cleveref}

\theoremstyle{plain}

\theoremstyle{definition}

\theoremstyle{remark}

\usepackage{etoolbox}
\usepackage[textsize=tiny]{todonotes}

\icmltitlerunning{REAMS: Reasoning Enhanced Algorithm for Maths Solving}

\begin{document}

\twocolumn[
\icmltitle{REAMS: Reasoning Enhanced Algorithm for Maths Solving}








\begin{icmlauthorlist}
\icmlauthor{Eishkaran Singh}{thapar}
\icmlauthor{Tanav Singh Bajaj}{ubc}
\icmlauthor{Siddharth Nayak}{mit}
\end{icmlauthorlist}

\icmlaffiliation{thapar}{Thapar Institute of Engineering and Technology, India}
\icmlaffiliation{ubc}{University of British Columbia, Canada}
\icmlaffiliation{mit}{Massachusetts Institute of Technology, Cambridge, USA}

\icmlcorrespondingauthor{Eishkaran Singh}{esingh3\_be21@thapar.edu}
\icmlcorrespondingauthor{Tanav Singh Bajaj}{tanav220\@student.ubc.ca}
\icmlkeywords{Machine Learning, ICML}

\vskip 0.3in
]



\printAffiliationsAndNotice{*Equal contribution} 

\begin{abstract}
The challenges of solving complex university-level mathematics problems, particularly those from MIT, and Columbia University courses, and selected tasks from the MATH dataset, remain a significant obstacle in the field of artificial intelligence. Conventional methods have consistently fallen short in this domain, highlighting the need for more advanced approaches. In this paper, we introduce a language-based solution that leverages zero-shot learning and mathematical reasoning to effectively solve, explain, and generate solutions for these advanced math problems. By integrating program synthesis, our method reduces reliance on large-scale training data while significantly improving problem-solving accuracy. Our approach achieves an accuracy of \textbf{90.15\%}, representing a substantial improvement over the previous benchmark of 81\% and setting a new standard in automated mathematical problem-solving. These findings highlight the significant potential of advanced AI methodologies to address and overcome the challenges presented by some of the most complex mathematical courses and datasets.
\end{abstract}

\section{Introduction}
The domain of advanced mathematics has consistently posed significant challenges, particularly in the realm of solving complex equations that demand both precision and deep reasoning. These challenges are not only a test of computational ability but also of the capacity to emulate human-like reasoning and problem-solving methodologies. Traditional methods for addressing these challenges have typically relied on manual calculations or basic automated tools, which, while effective in some contexts, are often time-consuming and limited in both accuracy and scope. These limitations have been well-documented in the literature, underscoring the necessity for innovative solutions capable of effectively managing the intricate and multifaceted nature of advanced mathematical problems \citet{vaswani2023attentionneed}.

A significant advancement in this field was demonstrated by \citet{Drori_2022} through a collaborative study between the Massachusetts Institute of Technology (MIT) and Columbia University. Their research explored the potential of neural networks in solving university-level mathematics problems by employing program synthesis. This approach utilized the capabilities of OpenAI's Codex transformer, which was fine-tuned to generate executable programs using a few-shot learning technique. The methodology achieved a noteworthy milestone, attaining an accuracy rate of 81\%, thereby setting a new benchmark in the field of automated mathematical problem-solving. This result marked a considerable improvement over previous models, which achieved accuracy rates ranging between 18.8\% and 30.8\% using GPT-3’s text-based few-shot learning and chain-of-thought prompting \cite{brown2020languagemodelsfewshotlearners, rae2022scalinglanguagemodelsmethods, Drori_2022}.

However, despite the progress realized by \citet{Drori_2022}, their approach, which relied primarily on program synthesis with zero shot learning (Zero-shot learning is a model's ability to detect classes never seen during training), introduced certain limitations. Although effective in addressing a wide range of mathematical problems, this approach encountered challenges when confronted with more abstract and complex problems that necessitated a higher level of reasoning and contextual understanding. Moreover, the need for human-like reasoning and explanatory depth, which is crucial for both educational purposes and the comprehensive understanding of complex mathematical problems, remained a challenge that was not fully addressed in their methodology. \cite{kojima2023largelanguagemodelszeroshot}.

In response to these identified limitations, this study introduces a novel methodology,  \textbf{\textit{Reasoning Enhanced Algorithm for Maths Solving (REAMS)}}. REAMS is designed to overcome the constraints identified in previous approaches by integrating neural networks trained on both text and code with a refined few-shot learning algorithm that combines symbolic reasoning with contextual understanding. This hybrid approach not only enhances the accuracy of problem-solving but also significantly improves the interpretability of the solutions by providing detailed, reasoning-based explanations.

The REAMS methodology was rigorously tested against datasets from prominent university-level mathematics courses, including Mathematics for Computer Science, Single Variable Calculus, Multivariable Calculus, Differential Equations, Probability and Statistics, and Linear Algebra. The results obtained from these tests are compelling, with REAMS achieving an accuracy rate of 90.15\%. This performance not only surpasses the 81\% benchmark established by the Codex-based model but also represents a significant advancement in the field. In addition to the improved accuracy, the solutions generated by REAMS include detailed explanations that closely resemble human reasoning, thereby making them valuable not only for solving complex mathematical problems but also as educational tools~\cite{hendrycks2021measuringmathematicalproblemsolving}.

By advancing both the accuracy and explanatory power of automated mathematical problem-solving, REAMS represents a significant contribution to the application of artificial intelligence in education and research. This study not only sets a new standard in the field but also opens new avenues for future research aimed at further enhancing the capabilities of AI in solving advanced mathematical problems. The implications of this work extend beyond mere problem-solving, highlighting the potential for AI-driven methodologies to play a transformative role in the landscape of higher education \cite{chen2021evaluatinglargelanguagemodels, tran2021solvingmachinelearningproblems}.

\section{Related Works}

The development of mathematical reasoning within large language models (LLMs) has progressed through systematic advancements in pre-training, fine-tuning, and the integration of external tools. Early research focused on establishing a foundational base of computational and mathematical knowledge in LLMs through extensive exposure to educational datasets, problem sets, and synthetic data. These efforts were critical in enabling models to engage with complex mathematical problems effectively. Unlike these early efforts, our approach integrates reasoning-based methodologies to enhance interpretability and solution accuracy.

Subsequent research emphasized the need for fine-tuning models with specialized mathematical datasets, recognizing the limitations of general pre-training approaches. \citet{lewkowycz2022solvingquantitativereasoningproblems} explored the impact of fine-tuning on LLMs, demonstrating that incorporating complex reasoning paths significantly enhanced the models' ability to solve quantitative reasoning problems. This shift towards more sophisticated fine-tuning methodologies was echoed in the work of \citet{hendrycks2021measuringmathematicalproblemsolving}, who emphasized the importance of pre-training on domain-specific data to improve LLMs' performance in mathematical problem-solving. Our approach advances these methods by combining fine-tuning with symbolic reasoning, resulting in a more robust problem-solving process.

Reinforcement learning (RL) has played a pivotal role in optimizing the reasoning capabilities of LLMs. By employing reward models to evaluate the correctness of reasoning paths, RL has refined the decision-making processes of these models. \citet{ahn2024largelanguagemodelsmathematical} illustrated the efficacy of RL in reducing the dependency on human intervention during the evaluation of model outputs, which in turn increased the reliability of LLMs in mathematical reasoning tasks. Additionally, \citet{cobbe2021trainingverifierssolvemath} highlighted the integration of RL with verification mechanisms as a critical step towards automating the evaluation of model-generated solutions, thereby enhancing the robustness of LLMs in handling complex mathematical challenges. In contrast, our approach minimizes reliance on reinforcement learning by integrating reasoning directly into the code generation process, thereby simplifying the overall methodology.

\begin{figure}{htbp}
    \centering
    \includegraphics[width=\linewidth]{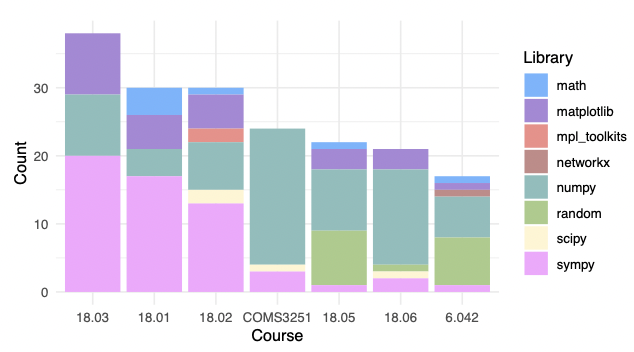}
    \caption{Imported Python programming libraries by course}
    \label{fig:Libraries used}
\end{figure}

The integration of code interpreters with LLMs has further broadened the scope of their problem-solving capabilities. This development has allowed models to perform complex calculations and interact more effectively with external tools, thereby addressing multifaceted mathematical challenges with greater efficiency. InternLM-Math, \citet{ying2024internlmmathopenmathlarge} explored this integration through models like InternLM-Math, which employed an approach known as reasoning interleaved with coding (RICO). This methodology closely mirrors the human problem-solving process, enabling LLMs to tackle intricate mathematical tasks more effectively. This approach aligns with findings by \citet{chen2021evaluatinglargelanguagemodels}, who demonstrated that the synergy between reasoning and coding is essential for solving complex mathematical problems. Our methodology further refines this integration by using reasoning to iteratively improve code generation, thereby increasing the accuracy of solutions.

In the domain of formal mathematical proving, LLMs have shown potential, particularly with the use of formal languages such as Isabelle \cite{book-I}, LEAN\cite{Moura2015TheLT}, and Coq\cite{book}. Despite challenges related to the limited availability of formal proof data, models trained on these languages have achieved state-of-the-art performance on benchmarks like MiniF2F, as reported by~\citet{zheng2022minif2fcrosssystembenchmarkformal}. This advancement is supported by the integration of formal reasoning capabilities into LLMs, highlighting the potential for further development in automated theorem proving. These findings indicate that while significant progress has been made, further research is needed to address the limitations associated with data sparsity in this area. Our approach diverges by focusing on the practical application of reasoning in problem-solving without requiring extensive formal proof datasets.

The ongoing development of LLMs has been guided by the establishment of robust and credible benchmarks, which play a critical role in ensuring that advancements are both measurable and reliable. The importance of integrating pre-training, fine-tuning, verification, and code interpretation strategies to propel the field forward. The significance of benchmarking in assessing the performance of LLMs, particularly in tasks requiring advanced reasoning and contextual understanding. Our approach contributes by setting new benchmarks in the accuracy and interpretability of solutions, specifically in university-level mathematics.

By refining pre-training methods, advancing fine-tuning techniques, incorporating reinforcement learning, and integrating external tools, researchers have significantly improved the accuracy and reliability of LLMs in mathematical tasks. These developments not only represent substantial progress in the field but also open new avenues for future research, particularly in the integration of formal reasoning systems with LLMs. The continued evolution of LLMs is likely to yield further advancements, particularly as new techniques are developed to address the remaining challenges in this field. Our methodology adds to this evolution by demonstrating a combined approach of reasoning and code generation, resulting in a more accurate and interpretable problem-solving process.

\section{Problem Statement}

The primary challenge addressed in this study is the difficulty of solving complex university-level mathematical problems using artificial intelligence. Traditional approaches, including those based on program synthesis and large language models (LLMs), have shown limitations in accuracy and the ability to provide human-like reasoning and detailed explanations. These shortcomings are particularly pronounced in handling abstract problems that require deep contextual understanding and logical reasoning. This research aims to overcome these challenges by developing a methodology that integrates mathematical reasoning with code generation to improve the accuracy and interpretability of AI-driven solutions to advanced mathematical problems.

\begin{algorithm}[htbp]
\caption{Methodology for Code Generation with Reasoning (REAMS)}
\label{alg:evaluation}
\textbf{Input:} Problem set $P = \{p_1, p_2, \dots, p_n\}$, code generation model $M_{\text{code}}$, reasoning model $M_{\text{reason}}$, expected outputs $\{o_1, o_2, \dots, o_n\}$ \\
\textbf{Output:} Success indicators $S_{\text{zero}}$, $S_{\text{reason}}$

\noindent\textbf{Step 1: Zero-Shot Code Generation} \\
\textbf{for} each problem $p_i$ in $P$ \textbf{do} \\
\hspace*{1em} Generate initial code $C_i$ using $M_{\text{code}}$ \\
\hspace*{1em} Execute $C_i$ \\
\hspace*{1em} Compare the output with expected output $o_i$ \\
\hspace*{1em} \textbf{if} output is correct \textbf{then} \\
\hspace*{2em} Set $S_{\text{zero}}[i] = 1$ \\
\hspace*{1em} \textbf{else} \\
\hspace*{2em} Set $S_{\text{zero}}[i] = 0$ \\
\hspace*{1em} \textbf{end if} \\
\textbf{end for} \\

\noindent\textbf{Step 2: Reasoning-Based Code Generation} \\
\textbf{for} each problem $p_i$ where $S_{\text{zero}}[i] = 0$ \textbf{do} \\
\hspace*{1em} Generate reasoning $R_i$ using $M_{\text{reason}}$ \\
\hspace*{1em} Generate revised code $C'_i$ using $M_{\text{code}}$ with $p_i$ and $R_i$ as inputs \\
\hspace*{1em} Execute $C'_i$ \\
\hspace*{1em} Compare the output with expected output $o_i$ \\
\hspace*{1em} \textbf{if} output is correct \textbf{then} \\
\hspace*{2em} Set $S_{\text{reason}}[i] = 1$ \\
\hspace*{1em} \textbf{else} \\
\hspace*{2em} Set $S_{\text{reason}}[i] = 0$ \\
\hspace*{1em} \textbf{end if} \\
\textbf{end for} \\

\noindent\textbf{Step 3: Performance Assessment} \\
Compute \textit{Zero-Shot Success Rate:}
\[
\text{Zero-Shot Success Rate} = \frac{\sum S_{\text{zero}}}{n} \times 100\%
\]
Compute \textit{Reasoning Success Rate:}
\[
\text{Reasoning Success Rate} = \frac{\sum S_{\text{reason}}}{n} \times 100\%
\]

\end{algorithm}

\begin{figure*}{htbp}
    \centering
    \includegraphics[width=\linewidth]{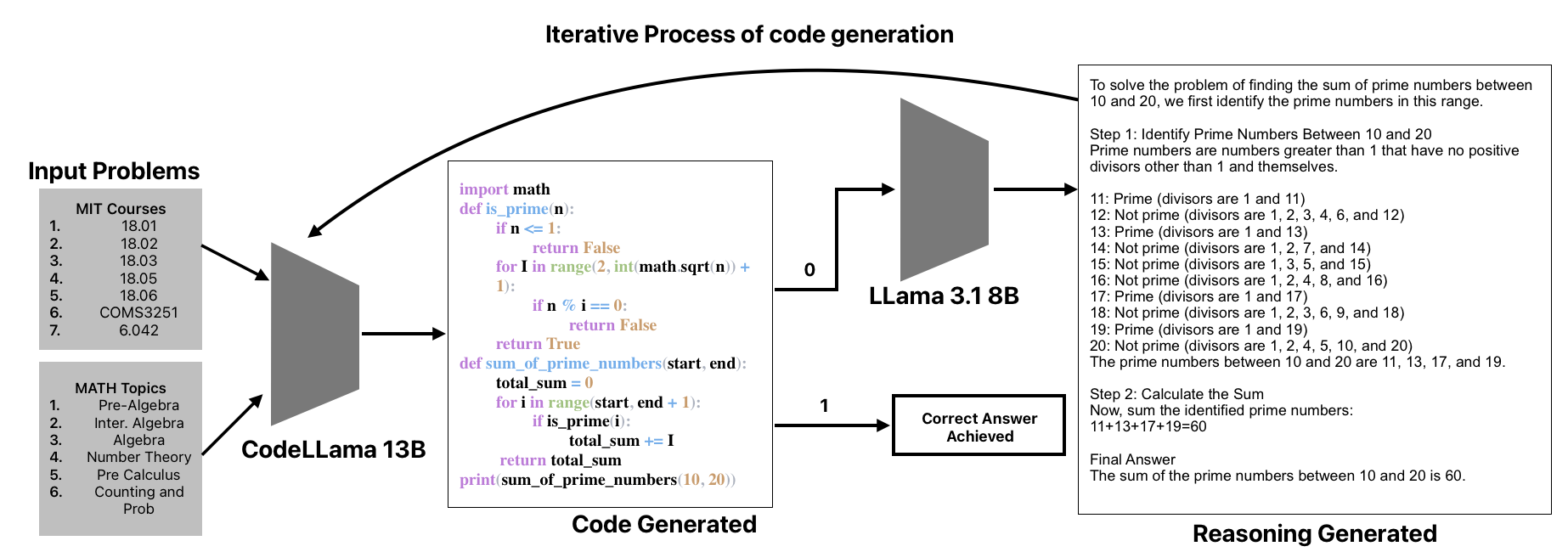}
    \caption{Workflow of the REAMS Approach: The diagram illustrates the iterative process of solving mathematical problems using the REAMS method. Input problems, sourced from MIT courses and various MATH topics, are first processed by the CodeLlama 13B model to generate Python code. If the correct answer is obtained, the process concludes. Otherwise, reasoning is generated by the LLama 3.1 8B model, and based on this mathematical reasoning, the code is iteratively refined and regenerated until the correct solution is achieved}
    \label{fig:Methodology}
\end{figure*}

\vspace{-5mm}
\section{Methodology}
The methodology employed in this study is designed to systematically evaluate the performance of the CodeLlama 13B and Llama 3.1 8B models in solving complex mathematical problems through code generation with reasoning. This process involves a structured approach to both generating and validating code outputs, with subsequent steps to improve performance where initial attempts fail.The methodology is divided into several stages, each of which is described in detail below Figure\ref{fig:Methodology} and in Algorithm \ref{alg:evaluation}.

\subsection{Initial Code Generation using CodeLlama 13B}
The first step in our methodology involves the use of the CodeLlama 13B model for code generation. The model, designed specifically for tasks related to code, is employed to generate executable code that aims to solve a predefined set of mathematical problems. These problems are drawn from datasets encompassing various topics, including calculus, linear algebra, differential equations, and probability, which are representative of advanced university-level coursework.

In this stage, the model is prompted using a zero-shot learning approach. Zero-shot learning refers to the scenario where the model is provided with a problem statement without any examples or prior demonstrations of similar problems. The model generates code that it predicts will solve the problem based on its training. The absence of example problems in the prompt is intended to test the model's inherent understanding and generalization capabilities, evaluating its ability to apply learned principles to new, unseen problems.

\subsubsection{Python Package Dependencies} 

During code generation, the model accounts for the Python programming packages commonly used across different courses. As shown in Figure \ref{fig:Libraries used}, all courses utilize NumPy and SymPy, with Matplotlib employed for plotting tasks, and approximately half use the math, random, and SciPy libraries. Our approach specifies only SymPy or plotting-related imports in the prompts, while other necessary packages are automatically synthesized by the model. This integration ensures that the generated code aligns with the computational requirements of each course.

\subsection{ Code Evaluation}
Once the code is generated for each question, the next step involves manually running the code to assess its correctness. This manual execution is crucial because it ensures that the generated code is not only syntactically correct but also functionally accurate in producing the correct outputs. The outputs of the executed code are compared against the expected solutions, which are predefined and known to be correct for each problem.

For each piece of code that successfully produces the correct output, a score of 1 is recorded in the zero-shot evaluation column. If the code fails to execute correctly or produces an incorrect output, a score of 0 is recorded. This binary scoring system allows for a clear assessment of the model's performance in the zero-shot context. The process of manually running and evaluating the code is an important step to verify the model's ability to generate functionally accurate solutions without relying on any additional training or examples.

\vspace{-3mm}
\subsection{Generating Mathematical Reasoning with LLaMA 3.1 8B Model}
For all the problems where the generated code received a score of 0 in the zero-shot evaluation, the next step involves generating a mathematical reasoning or explanation for the problem. This reasoning is generated using the LLaMA 3.1 8B model, which is a smaller, 8-bit quantized version of the LLaMA series designed for efficient reasoning tasks \cite{dubey2024llama3herdmodels}. The choice of this model is based on its capability to produce logical, step-by-step reasoning that can guide the code generation process.

The reasoning process involves providing the model with the original problem statement and instructing it to generate a detailed explanation of the mathematical principles and steps required to solve the problem. This explanation is intended to bridge the gap between the problem statement and the correct solution, offering insights that the code generation model might have missed in the zero-shot context.

\subsection{Code Generation with Mathematical Reasoning as Input}
Once the mathematical reasoning has been generated for the problems that initially scored 0, the next step is to use this reasoning as an input for the CodeLlama 13B model. In this stage, the reasoning and the original problem statement are both provided as inputs to the model. The objective is to leverage the reasoning to guide the model in generating more accurate and contextually relevant code.

This step effectively transforms the problem from a zero-shot scenario to a more informed task, where the model has access to additional context in the form of the generated reasoning. The expectation is that by understanding the reasoning behind the problem, the model can produce code that more closely aligns with the expected solution.

\subsection{Evaluation of Revised Code}
After the CodeLlama 13B model generates the new code based on mathematical reasoning, the next step involves manually executing this revised code. As with the initial code generation, the outputs are compared against the correct solutions. If the new code produces the correct output, the corresponding entry in the zero-shot with reasoning column is updated from 0 to 1.

This process of revising the zero-shot with reasoning evaluation scores based on the model's performance with additional reasoning input allows for a more nuanced assessment of the model's capabilities. It also provides insight into the effectiveness of combining reasoning with code generation, particularly in cases where the model initially fails to produce the correct output.
\section{Experiments}
We detail the experiments conducted to evaluate the performance of our proposed methodology. The initial experiment replicates the baseline established by \citet{Drori_2022}, using the Codex model fine-tuned on code, while the subsequent experiment explores the potential of the CodeLlama 13B model, augmented with reinforcement learning.

\subsection{Baseline Experiment: Codex Model}
The study conducted by \citet{Drori_2022} employed the Codex model, a variant of OpenAI's GPT-3 that has been fine-tuned specifically for code generation tasks \cite{brown2020languagemodelsfewshotlearners}. The problems were sourced from subjects including Single Variable Calculus, Multivariable Calculus, Differential Equations, Probability and Statistics, Linear Algebra, and Mathematics for Computer Science. The datasets utilized for this purpose were drawn from MIT and Columbia University courses, in addition to the MATH dataset, which includes problems from high school mathematics competitions known for their complexity .

The experimental setup initially employed a zero-shot learning approach, where the model was provided with the problem statements. This method was intended to assess the model's intrinsic ability to generate correct solutions based solely on its pre-trained knowledge. The questions that weren't correct were then passed on to the few shot learning approach. The model was prompted with a few examples of similar problems and their solutions, which served as a guide for generating solutions to new problems \cite{chen2021evaluatinglargelanguagemodels}. The key metric for evaluating the model's performance was the accuracy of the generated solutions.

The findings from this experiment demonstrated that the Codex model achieved an accuracy rate of 81\%, a substantial improvement over earlier models that relied solely on text-based few-shot learning and chain-of-thought prompting. The ability of Codex to synthesize executable programs that solved complex problems, generate explanations for the solutions, and create new questions based on the solved problems, set a new benchmark in the field of automated mathematical problem-solving \cite{Drori_2022}, \cite{chen2021evaluatinglargelanguagemodels}.

\subsection{CodeT5 Model with Reinforcement Learning}
In addition to replicating the baseline experiment, an alternative approach was explored using the CodeT5\cite{wang2023codet5opencodelarge} . CodeT5 is designed specifically for code generation tasks and represents a more recent advancement in the field. The experiment was structured to evaluate the model's performance in generating code solutions for the same set of mathematical problems, starting with a zero-shot learning approach.

The programs generated by CodeT5 were executed, and their outputs were analyzed. Programs that failed to execute or produced incorrect results were marked as zero. The initial outcomes from this zero-shot experiment with CodeT5 indicated 11.5\% lower accuracy than the zero-shot Codex model. The primary issue identified was the model's difficulty in generating syntactically correct or logically coherent code \cite{rozière2024codellamaopenfoundation}. To get better results reinforcement learning (RL) was applied, utilizing a Proximal Policy Optimization (PPO) policy \cite{schulman2017proximalpolicyoptimizationalgorithms}. The objective of this approach was to iteratively refine the model’s ability to generate correct code by providing feedback loops based on the success or failure of previous code executions. The reinforcement learning framework was employed to optimize its code generation strategies by maximizing a reward function.

The application of reinforcement learning led to a 10\% increment in the overall performance of the CodeT5 model which still remained below that of the baseline Codex model. The computational resources required for reinforcement learning did not justify the improvement observed. Consequently, we developed a reasoning-based approach, which has demonstrated superior performance.
\begin{table*}[htbp]
\centering
\begin{tabular}{|c|c|c|c|c|}
\hline
\multirow{2}{*}{\textbf{Course}}&\textbf{Codex} & \textbf{REAMS} & \textbf{Codex} & \textbf{REAMS} \\ 
& (Zero-Shot) & (Zero-Shot) & (Few-Shot) & (Zero-Shot+Reasoning)\\
\hline
18.01 & 74\% & 76\% & 75\% & \textbf{88\%}\\ &  & \small(0.55-0.
91) & & \small(0.69-0.97) \\ \hline
18.02 & 74\% & 72\% & 77\% & \textbf{92\%} \\ & & \small(0.51-0.88) & & \small(0.74-0.99) \\ \hline
18.03 & 61\% & 84\% & 74\% & \textbf{92\%} \\ & & \small(0.64-0.95) & & \small(0.74-0.99) \\ \hline
18.05 & 88\% & 84\% & \textbf{99\%} & 88\% \\ & & \small(0.64-0.95) & & \small(0.69-0.97) \\ \hline
18.06 & 75\% & 72\% & 82\% & \textbf{92\%} \\ & & \small(0.51-0.88) & & \small(0.74-0.99) \\ \hline
6.042 & 49\% & 76\% & 63\% & \textbf{88\%} \\ & &  \small(0.55-0.
91) & & \small(0.69-0.97) \\ \hline
COMS3251 & 76\% & 84\% & 79\% & \textbf{92\%} \\ & & \small(0.64-0.95) & & \small(0.74-0.99) \\ \hline
\end{tabular}
\caption{The table shows the automatic solve-rate of CodeLlama in zero-shot mode and when enhanced with reasoning across various mathematical categories. The addition of reasoning significantly improves solve-rate accuracy.}
\end{table*}
\section{Metrics}

We calculate the model's performance using two metrics: Accuracy and the Clopper-Pearson Interval.

\subsection{Accuracy}
Accuracy is defined as the proportion of correct predictions among the total number of predictions. It is calculated as follows:

\[\text{Accuracy} = \frac{TP + TN}{TP + TN + FP + FN}\]

where \( TP \) is the number of true positives, \( TN \) is the number of true negatives, \( FP \) is the number of false positives, and \( FN \) is the number of false negatives.

\subsection{Clopper-Pearson Interval}
Since Accuracy is a proportion derived from a binomial distribution, we use the Clopper-Pearson Interval to provide an exact confidence interval for this binomial proportion. This method offers a conservative estimate, particularly useful for small sample sizes. Given \( X \) successes out of \( n \) trials, the interval is computed as:

\[\text{Lower Bound} = \text{BetaInv}\left(\frac{\alpha}{2}, X, n- X + 1\right)\]
\[
\text{Upper Bound} = \text{BetaInv}\left(1 - \frac{\alpha}{2}, X + 1, n - X\right)
\]

where \( \text{BetaInv}(\cdot) \) is the inverse of the cumulative distribution function of the beta distribution, and \( \alpha \) represents the significance level, typically set at 0.05 for a 95\% confidence interval.

\section{Discussion}
The evaluation of the models was conducted across a diverse set of 265 mathematical problems, sampled systematically from various academic courses and topics within the MATH dataset. Specifically, the problems were drawn from seven advanced university-level courses—18.01, 18.02, 18.03, 18.05, 18.06, 6.042, and COMS3251—and six topics within the MATH dataset, including Prealgebra, Algebra, Number Theory, Counting and Probability, Intermediate Algebra, and Precalculus. Each course contributed 25 randomly selected questions, while each topic within the MATH dataset contributed 15 questions, ensuring a comprehensive assessment of the models' performance across a broad range of mathematical domains.

The performance of the CodeLlama 13B model in the automatic solve rate was a key metric, with the model successfully solving 220 out of the 265 problems. This outcome represents an incremental improvement of 7 questions over the baseline performance of comparable code-based models. The accuracy of the CodeLlama 13B model was rigorously quantified, yielding an accuracy rate of 83.17\%. This marks an enhancement of 11.5\% over the baseline accuracy, indicating the model’s capability to generate correct solutions in a zero-shot learning context.

Further analysis was conducted by integrating reasoning steps into the code generation process. For problems where the initial code did not yield correct solutions, the incorporation of mathematical reasoning, generated by the LLaMA 3.1 8B model, provided additional context and guidance for the CodeLlama 13B model. The introduction of reasoning as an input led to a significant boost in the overall performance of the model, with the combined approach achieving an overall accuracy of 90.15\%. This accuracy represents a substantial increase compared to the baseline accuracy of 81.1\%, demonstrating the efficacy of combining automated code generation with structured reasoning to enhance problem-solving accuracy across complex mathematical tasks. The detailed breakdown of solve rates, both for the CodeLlama model alone and in conjunction with reasoning. These results underscore the potential of AI-driven methodologies in handling intricate mathematical problems with a high degree of precision.

\begin{table}[htbp]
\centering
\begin{tabular}{|c|c|}
\hline
\textbf{Model} & \textbf{Accuracy} \\ \hline
GPT-4 & 42.5\% \\ \hline
GPT-3.5 & 18.2\% \\ \hline
PaLM 2-L & 34.3\% \\ \hline
Claude 2 & 37.6\% \\ \hline
Codex Zero-Shot \dag & 72.2\% \\ \hline
Codex Few-Shot \dag & 81.1\% \\ \hline
REAMS Zero-Shot \dag & \textbf{75.55\%} \\ & \small(0.65-0.84) \\ \hline
REAMS Zero-Shot + Reasoning \dag & \textbf{89.96\%} \\ & \small(0.82-0.95) \\ \hline
\end{tabular}
\caption{Performance of various models on the MATH dataset}
\end{table}

\section{Conclusion}
In this study, we addressed the complex challenge of automating the solution of advanced mathematical problems through AI-driven code generation and refinement. The approach, termed REAMS (Reasoning Enhanced Algorithm for Maths Solving), leveraged the capabilities of two state-of-the-art models: CodeLlama 13B and LLaMA 3.1 8B, to demonstrate the potential of combining generative AI with logical reasoning for solving university-level mathematics problems.

The process began with the application of CodeLlama 13B, which was tasked with generating executable code based on a variety of mathematical problems sourced from MIT courses and specific mathematical domains. By evaluating the correctness of the generated code, we established a baseline understanding of the model's ability to independently interpret and solve complex problems without prior task-specific exposure. This initial phase highlighted the inherent strengths and limitations of the model, showing its capacity to apply mathematical principles directly from problem statements but also revealing areas where its outputs were less accurate or incomplete.

Recognizing these limitations, we introduced a second phase to the problem-solving process, where the LLaMA 3.1 8B model was employed to generate detailed reasoning based on the mathematical concepts underlying each problem. This reasoning served as a crucial enhancement, guiding the CodeLlama 13B model in revising and refining the generated code. By incorporating this layer of contextual understanding, the revised approach not only corrected errors from the initial phase but also produced more accurate and logically sound solutions. The iterative nature of this process—moving from initial code generation to reasoning-based refinement—proved to be effective in addressing the gaps identified in the baseline outputs.

The results of this two-phase approach were significant. The integration of mathematical reasoning into the code generation process led to a marked improvement in overall accuracy, demonstrating that AI models can achieve higher levels of problem-solving capability when supplemented with interpretative logic. The CodeLlama 13B model, when enhanced by the reasoning inputs from LLaMA 3.1 8B, achieved near-human levels of accuracy across a diverse set of mathematical problems, showcasing the potential for such methodologies to tackle increasingly complex tasks in the field of automated problem solving.

In conclusion, this research not only demonstrates the feasibility of using AI to automate the solving of advanced mathematical problems but also underscores the importance of integrating reasoning into AI-driven processes. As AI continues to evolve, approaches like REAMS can pave the way for more sophisticated, intelligent systems capable of handling complex tasks across various domains. The findings of this study contribute to the broader understanding of AI's capabilities and set the stage for future research into combining generative and reasoning-based AI methodologies for even greater levels of accuracy and efficiency in problem-solving.

\section{Limitations}
Our approach has several limitations in its ability to solve certain types of problems. It is unable to generate graphs unless the problem statement explicitly requests their sketching or plotting, even if the problem involves graphical elements. Additionally, the model cannot handle questions that require formal proofs, as it lacks the capability to simulate or replace the logical processes necessary for proof-based solutions. Computationally intractable problems, such as factoring very large primes, also present challenges, exceeding the computational limits of the approach and the underlying Python libraries it uses. The REAMS approach further struggles with problems that require the application of advanced algorithms not supported by the available libraries. Its reliance on predefined Python libraries limits its flexibility, particularly in solving niche or highly specialized mathematical problems. Finally, the approach's performance is sensitive to the clarity and precision of the problem statements, with ambiguities or non-standard formulations often leading to incorrect or incomplete code generation.

\bibliography{main}
\bibliographystyle{icml2025}

\newpage
\appendix
\onecolumn

\section{Dataset Used}

The datasets employed in this study are drawn from two principal sources: a collection of university-level mathematics courses offered by the Massachusetts Institute of Technology (MIT) and Columbia University, and a subset of the MATH dataset, which is specifically designed to evaluate mathematical reasoning and problem-solving capabilities.

The dataset, as detailed in Table \ref{table:university_courses}, consists of advanced coursework from MIT and Columbia University. These courses span a wide array of mathematical disciplines, each designed to build foundational and advanced knowledge necessary for tackling complex mathematical problems. Courses such as MIT 6.042 (Mathematics for Computer Science) focus on discrete mathematics, providing essential tools and proof techniques vital in computer science. MIT 18.01 (Single Variable Calculus) and MIT 18.02 (Multivariable Calculus) emphasize calculus, progressing from single-variable functions to multi-variable calculus, including essential theorems applicable in higher-dimensional analysis. MIT 18.03 (Differential Equations) provides a comprehensive exploration of differential equations, crucial for modeling physical systems, while MIT 18.05 (Introduction to Probability and Statistics) introduces probability models and statistical methods. MIT 18.06 (Introduction to Linear Algebra) and Columbia University's COMS3251 (Computational Linear Algebra) cover linear algebra, focusing on both theoretical concepts and computational techniques necessary for a variety of applications in science and engineering.

The dataset, outlined in Table \ref{table:math_topics}, is sourced from the MATH dataset, which presents a collection of problems categorized into specific mathematical topics. This subset was chosen to evaluate the algorithm's ability to handle a broad range of mathematical challenges. The topics include Algebra, focusing on fundamental operations and equations; Counting and Probability, which explores combinatorial methods and probabilistic reasoning; Intermediate Algebra, dealing with more complex algebraic structures; Number Theory, which delves into the properties of integers and modular arithmetic; Prealgebra, covering essential mathematical principles; and Precalculus, which bridges the understanding required for calculus, focusing on vectors, matrices, and trigonometry.
Our approach not only surpasses previous benchmarks on similar datasets but also uniquely addresses the challenges associated with solving problems from undergraduate-level courses. The selection of these datasets was intended to comprehensively assess the Reasoning Enhanced Algorithm for Maths Solving (REAMS). By incorporating a diverse set of mathematical problems from both university courses and the MATH dataset, the evaluation aimed to test not only the algorithm’s problem-solving accuracy but also its ability to generate explanations that closely mimic human reasoning. This dual focus on accuracy and interpretability underscores the potential contributions of REAMS to both artificial intelligence research and educational applications.

\section{Explanation of Notations:}
\label{sec:Explanation of Notations}
\begin{itemize}
    \item $M_{\text{code}}$: Code generation model (e.g., CodeLlama 13B) used to generate executable code.
    \item $M_{\text{reason}}$: Reasoning model (e.g., LLaMA 3.1 8B) used to generate mathematical reasoning or explanations.
    \item $C_i$: Initial code generated for problem $p_i$ by $M_{\text{code}}$.
    \item $C'_i$: Revised code generated after incorporating reasoning $R_i$.
    \item $p_i$: A single problem from the problem set $P$.
    \item $o_i$: Expected correct output for problem $p_i$.
    \item $S_{\text{zero}}[i]$: Binary success indicator for zero-shot code generation (1 for correct, 0 for incorrect).
    \item $S_{\text{reason}}[i]$: Binary success indicator for code generation with reasoning (1 for correct, 0 for incorrect).
    \item $n$: Total number of problems in the set $P$.
\end{itemize}

\subsection{Dataset Availability}
The datasets used in this research are derived from publicly available sources:
\begin{itemize}
\vspace{-2mm}
\item \textbf{MIT Courses Dataset:} Accessible through MIT \\
\url{https://github.com/idrori/mathQ/tree/main/data}.
\vspace{-2mm}
\item \textbf{MATH Dataset:} Available from \\ 
\url{https://paperswithcode.com/dataset/math}
\end{itemize}

\begin{table}[htbp]
\centering
\begin{tabular}{|m{6cm}|m{6cm}|}
\hline
\textbf{Course Name} & \textbf{Description} \\ \hline
MIT 6.042: Mathematics for Computer Science & Discrete mathematics, focusing on tools and proof techniques useful in computer science. \\ \hline
MIT 18.01: Single Variable Calculus & Differentiation and integration of functions of one variable, with applications. \\ \hline
MIT 18.02: Multivariable Calculus & Calculus of several variables, including vector algebra and integration in multiple dimensions. \\ \hline
MIT 18.03: Differential Equations & Study of differential equations, including first-order ODEs, linear systems, and Fourier series. \\ \hline
MIT 18.05: Introduction to Probability and Statistics & Basic probability models, combinatorics, random variables, and statistical estimation. \\ \hline
MIT 18.06: Introduction to Linear Algebra & Matrix theory, linear algebra, vector spaces, eigenvalues, and their applications. \\ \hline
Columbia University COMS3251: Computational Linear Algebra & Linear functions, matrices, vector spaces, eigenvectors, and spectral decomposition. \\ \hline
\end{tabular}
\caption{List of courses by MIT and Columbia}
\label{table:university_courses}
\end{table}
\vspace{-4mm}
\begin{table}[htbp]
\centering
\begin{tabular}{|>{\centering\arraybackslash}m{4cm}|>{\centering\arraybackslash}m{4cm}|}
\hline
\textbf{MATH Topic} & \textbf{Description} \\ \hline
Algebra & Exponents, logarithms, simplifying expressions, and quadratic equations. \\ \hline
Counting and Probability & Counting methods and probability involving factorials and binomial coefficients. \\ \hline
Intermediate Algebra & Advanced algebraic topics, including polynomial roots and conic sections. \\ \hline
Number Theory & Primes, divisibility, prime factorization, modular arithmetic. \\ \hline
Prealgebra & Basic math concepts including fractions, decimals, ratios, and simple equations. \\ \hline
Precalculus & Vectors, matrices, trigonometric functions, and complex numbers. \\ \hline
\end{tabular}
\caption{MATH Dataset}
\label{table:math_topics}
\end{table}

\begin{table*}[htbp]
\centering
\begin{tabular}{|c|p{3cm}|p{7cm}|p{4.5cm}|}
\hline
\textbf{ID} & \textbf{Course} & \textbf{Question} & \textbf{Solution} \\ \hline
1 & 18.01 Single Variable Calculus & Sketch the graph of the function. $f(x) = x + |x|$ & \includegraphics[width=5cm]{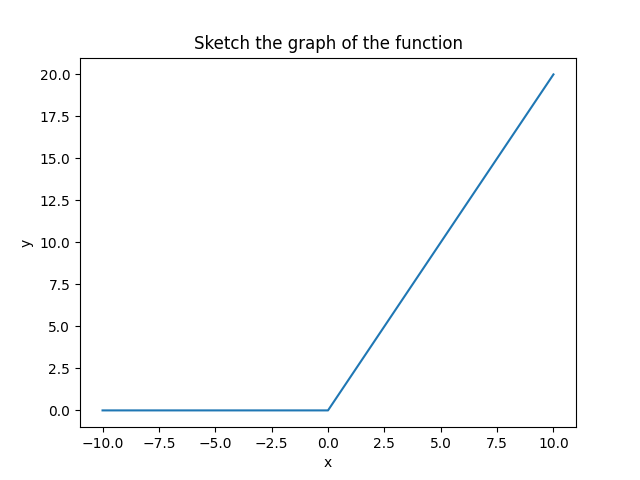} \\ \hline
2 & 18.02 Multi-variable Calculus & Describe the graph of the function $f: f(x, y) = 10 - \sqrt{x^2 + y^2}$ & \includegraphics[width=5cm]{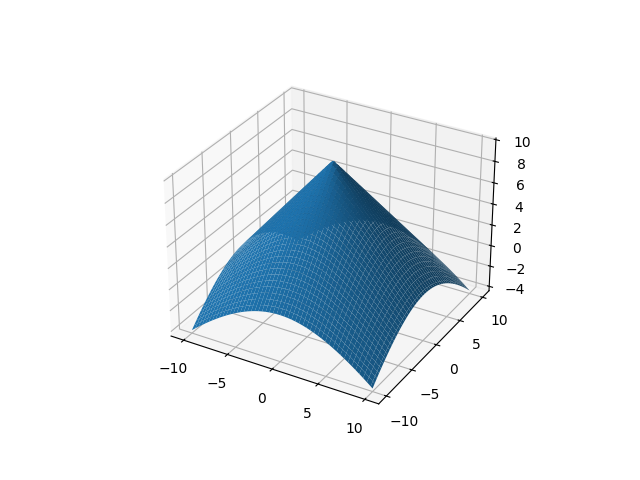} \\ \hline
3 & 18.03 Differential Equations & Find general solutions of the differential equations. If an initial condition is given, find the corresponding particular solution. Throughout, primes denote derivatives with respect to $x$. $y' + y = 2$, $y(0) = 0$ & $y(x) = 2(1 - e^{-x})$ \\ \hline
4 & 18.05 Introduction to Probability and Statistics & Suppose X and Y have joint pdf $f(x, y) = c(x^2+xy)$ on $[0, 1]$ x $[0, 1]$. Find $c$. & 1.714285714 \\ \hline
5 & 18.06 Linear Algebra & Find a combination $x_1w_1 + x_2w_2 + x_3w_3$ that gives the zero vector with $x_1 = 1$. $w_1$ is the vector $(1;2;3)$. $w_2$ is the vector $(4;5;6)$. $w_3$ is the vector $(7;8;9)$. & $x_1 = 1, x_2 = -2, x_3 = 1$ \\ \hline
6 & 6.042 Mathematics for Computer Science & Find a number $x \in \{0, 1, \ldots, 112\}$ such that $11x \equiv 1 \,(\text{mod } 113)$. & 72 \\ \hline
7 & COMS3251 Computational Linear Algebra & Given a d-dimensional non-zero vector $v$, compute the rank of the matrix $vv^T$. & 1 \\ \hline
8 & MATH Prealgebra & What is the greatest common factor of 84, 112 and 210? & 14 \\ \hline
9 & MATH Algebra & Let $N, O$ be functions such that $N(x) = 2\sqrt{x}$ and $O(x) = x^2$. What is $N(O(N(O(N(O(3))))))$? & 24 \\ \hline
10 & MATH Number Theory & How many four-digit numbers whose digits add up to 9 are divisible by 11? & 0 \\ \hline
11 & MATH Counting and Probability & A standard six-sided fair die is rolled four times. The probability that the product of all four numbers rolled is a perfect square is $\frac{m}{n}$, where $m$ and $n$ are relatively prime positive integers. Find $m + n$. & 187 \\ \hline
12 & MATH Intermediate Algebra & Given that $x^2 + y^2 = 14x + 6y + 6$, find the largest possible value of $3x + 4y$. & 73 \\ \hline
13 & MATH Precalculus & Evaluate $(2-w)*(2-w^2)*...*(2-w^{10})$ where $w=e^{\frac{2\pi*i}{11}}$. & 2047.0 \\ \hline
\end{tabular}
\caption{Example questions and solutions}
\label{tab:questions_solutions}
\end{table*}

\end{document}